\documentclass[10pt, a4paper]{article}

\usepackage[final]{lrec2026} 

\usepackage[utf8]{inputenc} 
\usepackage[T1]{fontenc}    
\usepackage{hyperref}       
\usepackage{url}            
\usepackage{booktabs}       
\usepackage{amsmath}        
\usepackage{amsthm}
\usepackage{amsfonts}       
\usepackage{nicefrac}       
\usepackage{microtype}      
\usepackage{xcolor}         
\usepackage{multirow}       
\usepackage{rotating}

\newtheorem*{definition}{Definition}

\title{The Case for Repeatable, Open, and Expert-Grounded Hallucination Benchmarks in Large Language Models}

\name{Justin D. Norman, Michael U. Rivera, D. Alex Hughes}

\address{
    University of California, Berkeley, School of Information \\
         \{justin.norman, michaelrivera, dhughes\}@berkeley.edu}

\abstract{
 Plausible, but inaccurate, tokens in model-generated text are widely believed to be pervasive and problematic for the responsible adoption of language models. Despite this concern, there is little scientific work that attempts to measure the prevalence of language model hallucination in a comprehensive way. In this paper, we argue that language models should be evaluated using repeatable, open, and domain-contextualized hallucination benchmarking. We present a taxonomy of hallucinations alongside a case study that demonstrates that when experts are absent from the early stages of data creation, the resulting hallucination metrics lack validity and practical utility.
 \\ \newline  }

\begin{document}

\maketitleabstract

\section{Introduction}

The current landscape of hallucination benchmarking is hampering progress toward the development of models that perform efficiently and produce fewer errors and hallucinations. Because existing benchmarks are (a) not repeatable, (b) not open, and (c) either domain-agnostic or isolate expertize behind private walls, researchers and commercial developers cannot openly collaborate and compare model performance in the pursuit of faster, more capable, and less-carbon intensive frameworks. Developing open, repeatable benchmarks would enable the language modeling community to establish a shared foundation for model training and experimentation prior to submitting models to public evaluation. When these benchmarks incorporate expert knowledge, they help ensure that resulting models are both valid and practically useful.

This paper proceeds as follows. We begin with an outline of the problem space of hallucination benchmarking. Next, we summarize the state of the field, provide a working definition of model hallucination, and illustrate the real-world consequences of hallucination through two examples. We then argue that a repeatable, open, domain-contextualized approach will produce fewer hallucinations. Next, we demonstrate how expert responses differ from model-generated text, drawing on a case study from medicine and another from politics. We conclude by highlighting some remaining challenges in the field. 

A crucial first step in identifying and mitigating hallucinations in contemporary Large language models (LLMs) is the development of a process to create a golden dataset that supports domain-specific hallucination benchmarks. One key lesson we learned---the hard way---is that domain expertise must be present from the very inception of the dataset. Without the involvement of domain experts in the early stages of data creation, hallucination metrics risk lacking both scientific validity and real-world utility. 

\section{Language Model Hallucination}

Advances in open and repeatable benchmarks \citep[e.g.,][]{srivastava2023, bbeh2025} have helped to constrain general-purpose, model-generated text resulting in text that has become increasingly, though only seemingly, coherent. But, what was initially regarded as an impressive novelty is not being incorporated into academic, civil-society, and corporate processes. Language models are being pushed to produce outputs that serve concrete business, policy, medical and scientific purposes and stakeholders are increasingly relying on language models to take expert-like actions in high-stakes domains. This creates an acute problem: it is very unlikely that current models are ready to operate at the level of rigor or reliability that is expected from a domain expert. Although it is generally known that recent model vintages are producing hallucinations \cite{Zhou2024larger_and_more, kalai2025languagemodelshallucinate}, the implications of the widespread implementation of these models is less widely discussed. Indeed, although there is remarkable enthusiasm to use language models, there is virtually zero understanding of how these models perform at expert and near-expert tasks. 

Early efforts in LLM hallucination benchmarking primarily centered on summarization tasks that compared model-generated summaries against reference texts to assess factual consistency (e.g., CNN and DailyMail \citep{hermann2015, narayan2018} and XSum \citep{narayan2018}). As LLMs evolved and were used to generate seemingly novel text, new benchmarks that did not rely upon scoring against reference texts became necessary. When benchmarking metrics scored only perceived coherence or helpfulness, teams' trained models to score better on these measures, but with the side effect of wild hallucinations---entirely ungrounded references, links, and factual claims. 
Recent advancements have moved from evaluating the perceived coherence of language, to assessing the factual accuracy of model-generated text in zero-shot question answering \citep[see, e.g.,][]{Hughes_Vectara_Hallucination_Leaderboard_2023,  li2023haluevallargescalehallucinationevaluation}. 

We highlight two recent advancements in hallucination detection: the Vectara hallucination leaderboard and HaluEval. The Vectara hallucination leaderboard provides an open and continuously updated platform for evaluating hallucination rates across LLMs using document summary metric, the Hughes Hallucination Evaluation Model (HHEM) \citep{Hughes_Vectara_Hallucination_Leaderboard_2023, hhem-2.1-open}. HHEM assesses factuality using a zero-shot question-answering setup over grounded documents; humans define ground-truth, binary judgments of correctness. In a separate project, HaluEval collects 45,000 human-annotated samples across multiple generation tasks (e.g., question answering, summarization, open-domain generation), distinguishing between extractive, abstractive, and factual hallucinations \citep{li2023haluevallargescalehallucinationevaluation}. 
Crucially, HaluEval 2.0,  provides a domain-relevant benchmarks across biomedicine, finance, science, education, and as well as an open category \citep{li2024dawndarkempiricalstudy}. 

Despite these advancements, we remark that the field has focused on only a small subset of possible \textit{dimensions} from the full range available to evaluate and constrain models. These selected dimensions include tasks like causal reasoning, geometric rotation, and calendar scheduling under conditions of incomplete information \citep[see, e.g.][]{bbeh2025}. Certainly these are difficult tasks for machines that are trivial for humans; but current implementations do not focus on calendar scheduling tasks. Said simply, existing hallucination benchmarks are not informed by the type of expertise that is necessary to constrain models working in hard contexts, and so hallucination continues to be a challenge for models working on expert-level tasks. 

\subsection{Defining LLM Hallucination}

While there is a considerable body of work surrounding language model hallucination, the field has not settled on a single operational definition of what, \textit{specifically} constitutes a hallucination \citep[see e.g.,][for a survey of the field]{ziwei2023, huang2025survey}. The term hallucination arises in 2017 in the context of a data-to-text generation task \citep{wiseman2017challenges}, features in abstractive summarization in 2020 \citep{maynez2020faithfulness}, and features prominently in \textit{Stochastic Parrots} \citep{bender2021dangers}. To create a point of reference within this paper, we define language model hallucinations in the following way:

\begin{definition} 
    \label{def:hallucination}
    An LLM hallucination is a generated text statement presented without qualifying language that is: incoherent, factually incorrect, or unverified in the judgment of a human expert capable of making that assessment.
\end{definition} 

\paragraph{Qualifying Language} \textit{Qualifying language} is language that conveys either the probabilistic nature of the output or the lack of uncertainty about the true state of the world. A response that does not include such language makes a seemingly authoritative statement or fails to acknowledge the existence of alternatives.  

\paragraph{Incoherence}
Current generation language models produce syntactically correct, seemingly coherent strings of text \citep{bender2021dangers}. Indeed, it was the initial emergence of seeming coherence, paired with an engaging user-facing design that led to the eruption of popular attention to language models in 2022. Bender and Gebru coin the evocative term ``stochastic parrots'' to highlight the risks of this seemingly coherent language, writing, ``Our perception of natural language text, regardless of how it was generated, is mediated by our own linguistic competence and our \textit{predisposition to interpret communicative acts as conveying coherent meaning and intent}'' \citep[][p. 616]{bender2021dangers}. In other words, because model generated text reads like language that a human expert might produce, humans have a blind-spot that leads us to interpret the text as possessing meaning, intent, and as we argue, expertise. First generation, non-expert-derived benchmarks have largely solved issues of coherence and model generated text is nearly uniformly coherent on general-language tasks.

\paragraph{Factually Incorrect}
LLM text that misstates a verifiable fact is currently the most crucial form of hallucination to address for near-expert performance. Working on document summarization, Huang et. al. identity two types of fact-based hallucinations: \textit{(1) intrinsic  errors} where text presents statements that can be verified against known corpora; and \textit{(2) extrinsic errors} where models make up facts to support statements, but these facts are inconsistent with verifiable real-world knowledge. Importantly, when working on expert-like tasks, both intrinsic and extrinsic errors require expertise to evaluate whether model generated text is correct or hallucinated. Extrinsic errors can be controlled in two ways: either (a) creating tighter safeguards and guardrails that cause models to more readily default to ``I don't know'' behavior; or (b) providing more expertise. Presently, factual contradiction is undertaken by selecting only ``high-quality'' data sources (see Lucy et. al. for important context surrounding data filtering \citep{lucy2024aboutme}), Retrieval Augmented Generation (RAG), few-shot training and knowledge graph based approaches, and, recently agent-based approaches. The profusion of approaches belies the reality that this type of factual incorrectness is the most pressing current problem, and so has been the locus of considerable development.  

\paragraph{Unverified}
Unverified responses that are those that either fail to cite a source, or cite an external source that is insufficient to corroborate the claim --- whether to confirm or contradict it \citep{ziwei2023} \citep{huang2025survey}. For example, if a user asks about eligibility requirements to vote in California, an LLM hallucination may either fail to cite a source altogether or cite one --- such as a nonprofit organization or the California Secretary of State --- where the source cited is insufficient to verify the validity of the LLM response.

\subsection{Consequences of Language Model Hallucination}

Language model hallucination is not simply a problem to be solved so that models score better on benchmarks. These models power widely used consumer products, and when users read and act on hallucinated output, there is a real risk of harmful consequences. Because communication is so core to the human experience, syntactically correct language serves as a credibility hack \citep{chomsky1976, christiansen2008}, effectively jailbreaking our perceptive system to imbue what seems like coherence from a stochastic parrot \citep{bender2021dangers}.  LLM models can hide their lack of coherence within the written word --- LLMs produce text that impersonates the writing of experts, and are imbued with ``expertise'' by readers. When combined with human decision making which is well-known to be prone to using lower-cost heuristic evaluation \citep{simon1955, simon1956, kahneman2011} --- even among high-stakes decision makers with very strong financial and political incentives for decision accuracy \citep{hafner2013} --- this jailbreak creates a profound risk for misinterpretation, and evaluation of model-generated language \citep{bender2021dangers}. In this section, we introduce two examples which we present to domain experts.

\paragraph{Voters Seeking Information on Political Process} 
The manner for how elections are conducted is constitutionally devolved to the states (Article 1, Section 4, US Constitution). Although federal law stipulates an common election day, states are largely responsible for determining the means and methods for how elections are run. Consequently, if a voter were to move from one state to another, it is possible that the election process might look quite different. The differences may include: which state agency handles registration; registration requirements and dates; availability of absentee or early voting; voter identification requirements; among others. In this context, it would be quite reasonable for a voter to query an LLM with a question like, ``What documentation do I need to vote, and when does early voting start in [my location]?'' Queries of this type generated documented instances of hallucination that would lead people to the wrong early voting date. These hallucinations persisted even when the model was instructed to describe the ``click-path'' that was used to navigate from an official government source to the specific resource used to produce an election date; the model generated text that traversed to a non-existent series of sites \citep{hughes2025}. For some period of time ChatGPT 3.5-x returned a caution to these political information queries, encouraging users to confirm information with their local elections agencies; however, these cautions are no longer present for users who are using ChatGPT 4o in April 2025.\footnote{OpenAI is also aware of the limitations of ChatGPT for expert users in the political space. In 2024, the company prohibited politicians and lobbyists from using ChatGPT for official campaign activities\citep{NPR2024}. There is also evidence that LLMs generated inaccurate and misleading responses for voters during the 2024 U.S. presidential primary \citep{nyt2025}.}

\paragraph{Medical Diagnosis}
Medical diagnosis is a challenging task. Patients noisily introspect their own state, commonly exhibit social desirability bias, and frequently present with comorbid diseases that are caused by a complex relationship between genetics, lifestyle, and proximate causes. Front line clinicians develop expertise that incorporates medical science and real-world clinical training and use this expertise to ask winnowing questions and to order informative tests to arrive at an manageable set of possible diagnoses. With this set, doctors propose medical interventions and observe how patients respond.

Compared to clinicians, language models are missing huge swaths of the diagnostic toolkit. Models do not have access to long clinical histories or family or social context; nor do they have the ability to manipulate, either physically to check the severity of an ankle sprain, or through prescription to see whether a fever responds to acetaminophen; nor can models order labs or imaging to distinguish between hidden states that are unknowable to a patient.\footnote{In spite of this, it is likely that people view LLMs as  substitutes for visits to the doctor's office. 
}
It is little surprise, then, that  model-based diagnostic accuracy is substantially lower than diagnostic accuracy of expert doctors \citep{takita2025systematic}. To the extent that existing work includes medical expertise in the creation and benchmarking of language models, they are largely ``bolting-on'' doctors' expert review at the end of the diagnostic pipeline \citep[see, e.g.,][]{wu2024llmsciterelevantmedical, kim2025}.

\section{Solution Space}
Here, we present the requirements of a scientific solution to address model hallucination. We build from the basic requirement of internal evaluation using repeatable tests; to the open sharing of tests and test performance to enable replication between research teams; and finally to our core argument that benchmarks need to contain carefully solicited expertise from experts in the domains where the model will be deployed. We present arguments for each of these three objectives separately, but would like to be quite clear that none of these are sufficient to serve as a whole benchmark. Instead, each is necessary to produce a functioning benchmark, and only when all three are objectives are implemented in concert can a benchmarking system effectively guide model development.

\begin{table*}[ht]
\centering
\small
\caption{Expert vs. Language Model Response Comparison. The \colorbox{yellow!30}{yellow text} denotes correct expert information, and the \colorbox{green!30}{green text} denotes hallucinated LLM content.}
\begin{tabular}{lp{12.5cm}}
\toprule
\textbf{Field} & \textbf{Content} \\
\midrule
\textbf{Prompt} & I've got frequent nosebleeds, bruising easily, and fatigue. Do I have aplastic anemia? \\
\midrule
\textbf{Expert Answer} & No, you might have \colorbox{yellow!30}{\parbox{9.9cm}{idiopathic thrombocytopenia, von Willebrand disease, malignancy}} \\
\midrule
\textbf{LLM Answer} & \colorbox{green!30}{\parbox{12.5cm}{Aplastic anemia is a rare condition in which the body stops producing enough new blood cells. Symptoms can include frequent nosebleeds, easy bruising, fatigue, and other signs of low blood cell counts such as paleness and shortness of breath.}} However, these symptoms can also be caused by other conditions. It is important to see a healthcare provider for a proper evaluation and diagnosis. \\
\midrule
\textbf{Hallucination} & Yes - Factual Contradiction and Factual Generalization \\
\midrule
\textbf{Hallucination Metrics} & $FCD^{+}=3.57,\ FGR^{+}=1.0,\ SCD^{-}=1.22,\ THS^{-}=2.57$ \\
\midrule
\textbf{Impact} & Expert explicitly rules out aplastic anemia and provides specific alternative diagnoses. LLM confirms aplastic anemia as possibility, missing critical differential diagnoses. METEOR$^{+}$=0.10 shows poor alignment with expert guidance. \\
\bottomrule
\end{tabular}
\label{tab:comparison_example3}
\end{table*}

\paragraph{Repeatable}
Language model developers should provide repeatable benchmarks that allow for internal experimentation and evaluation prior to submitting models for public benchmarking. Without repeatable internal benchmarks, teams must repeatedly submit against high-visibility, public evaluation benchmarks \citep[e.g.,][]{Hughes_Vectara_Hallucination_Leaderboard_2023, li2023haluevallargescalehallucinationevaluation, hong2024hallucinationsleaderboardopen, bbeh2025} creating perverse consequences for the advancement of the science of language models. 

When consumers choose model adoption based on highly-visible leader-boards, scientific and profit incentives are muddied. Recall, for example, that in 2015 a research team claimed a 10.2\% relative improvement over the state of the art \citep{ilsvrc2015announcement}. As would come to light through several revisions to the paper and considerable public hand-wringing by the contest organizers , the research team had submitted 200 distinct models to the scoring dataset, more than all other entrants combined \citep{simonite2015}. If development teams do not have repeatable benchmarks that can be used while building and training models, and instead have to evaluate their work on public scoring benchmarks, this problem will assuredly continue to arise \citep{wiggers2025crowdsourced}. Accelerating this problem are additional selection pressures: developers want to work on competitive teams, and technical decision makers want to purchase an implement the best performing models. 


\paragraph{Openness} 
Trust and repeatability are principal tenants of scientific investigation. As a field, computer science has a long history of building computing tools in the pursuit of this openness \cite[e.g.,][]{knuth1984texbook}. Not only does openness allow for the development and transmission of ideas between researchers, but it also allows for open evaluation and repeatable execution of tests conducted against data \citep{kluyver2016jupyter}. The open science movement that makes compute environments, scientific code, and data widely available has become \textit{de rigueur} for publication across many scientific communities, speeding discovery and error correction, and reducing malfeasance \citep{christensen2019}. Benchmarking openness, as we intend it, combines aspects of scientific openness and open source software (OSS). 

These principles are evident when write tests for students as instructors or teachers: We ask teaching assistants or graduate student instructors who are working in that class to write practice tests and test questions. Crafting these test questions often involve reviewing and refining those draft questions before they are actually made to be student facing. Under this model, the experts are contributing to both the practice sets (i.e. the public development and the actual test (i.e. the private evaluation sets) and we are relying on them to keep from sharing those answers with students. Current benchmarking frameworks make the draft sets public and benchmarking sets private, but do not have a mechanism for open contributions to either of these sources of ground truth. What we propose is not to break the hidden nature of the benchmark evaluation set, but rather that experts can contribute to the efforts.

Consistent with scientific openness, benchmark openness should be open about the benchmark questions, the benchmark code, and the benchmark scores. This type of scientific openness supports the repeatable benchmarking position presented in the last subsection. When consumers are exposed to technology products that are mere days off of the scientific development floor, these products must necessarily contain aspects of scientific evaluation. Openness about how a model performs against different benchmarks is a minimum ``safety evaluation'' that is due to consumers who are going to use a product. Just as medication labels contain statements of side effects, model publishers should describe the expected behaviors and error rates so that users can make minimally informed decisions \citep{mitchell2019model}. Consistent with the OSS movement, open benchmarks allow for distributed contributions from domain-experts ensuring benchmarks have broad topical coverage. What is more, open contributions distribute the maintenance costs for this evaluation tooling.

\paragraph{Domain Contextualized}
The state of the art benchmarking for LLMs evaluates models on tasks that are  challenging, yet are very narrow. General reasoning benchmarks evaluate reasoning capabilities across a range of broad tasks such adjudicating boardgames rule questions \citep{kazemi2024}, complex geometry problems \citep{kazemi2023} and causal reasoning \citep{bbeh2025}. Developers have chosen to drive models toward performance on these narrowly defined tasks with the consequence that the models hallucinate at higher rates in other zero- and few-shot prompts \citep{Zhou2024larger_and_more, nyt2025}. 

Simply stated, it is not possible for teams to develop effective benchmark evaluations in areas where they are non-expert. These teams lack knowledge of the key dimensions to consider for a topic outside their expertise as well as the skill to design questions that would evaluate evaluate models on those dimensions. These teams are able to write benchmarks that are difficult in their domain --- solvable by the average humans or skilled language model researchers --- but they fall short in representing deep knowledge in other fields. The result is that the best current community-generated, contextualized evaluation frameworks focus on reasoning tasks that do not generalize well to novel, domain-specific tasks. When models are developed without human expertise as a part of the training, they hallucinate more frequently when asked for expert-level advice compared to prompts that do not require expertise \citep{Zhou2024larger_and_more}.  

\section{Case Study in Hallucination Dataset Development}

In order to produce models that are both scientifically valid and useful in real-world settings, we develop a process that facilitates the comparison between expert, human generated text and non-expert, model-generated text. 

\paragraph{Question Development} 
We first develop sixty-four questions related to political procedures and medical diagnoses. For the political domain, we collaborated with the political practitioner to craft both the questions (e.g., ``Can I register to vote on election day in California?'' and ``Do I need ID to vote in Arizona?'') and the corresponding factual, verifiable answers about electoral procedures \citep[see e.g.,][]{ncsl2025election}. In the medical domain, our research team generated questions (e.g., ``I have ulcerative colitis. Will mesalamine help control it?'' and ``I’m experiencing joint pain, morning stiffness, and swollen fingers. Do I have psoriatic arthritis?'') using publicly available resources, and a physician provided plausible diagnoses. 

\paragraph{Expert, Ground Truth Answers}
We draw on the expertise of one of the authors, a political consultant with a PhD in political science, and a board-certified physician to develop our golden dataset in the domains of politics and medicine. As we discuss shortly, the way we engaged domain experts directly influenced the validity and usefulness of the resulting hallucination metrics.

We take responses provided by domain experts to be \textit{ground truth} (GT). How GT is conceptualized in each domain reflects real-world complexities and highlights our main argument --- domain-specific context is critical for the development of hallucination metrics. In the political domain, questions such as ``Who can register to vote in Arizona?'' have a single, correct answer that is defined by law. In contrast, in medicine, clinicians are trained to iteratively narrow toward a diagnosis based on patients' chief complaints, test results, and responses to medical intervention. Often, prior to intervention, there isn't a single, definitive conclusion. Ideally for performance benchmarking, GT would reflect a single state; however, we believe the GT should reflect the real-world as not-fully-known as it may be.

\paragraph{LLM Generated Answers}

To efficiently capture comprehensive responses across a variety of LLM types, sizes, providers and vintages, we developed an agent. Given a list of domain-specific perturbed prompts, the agent automatically queries a user-specified cohort of LLMs, captures the responses and associated metadata, and organizes the outputs in a JSON-formatted data structure for future retrieval and evaluation.

For the purposes of this paper, to establish a baseline of consistently hallucinated LLM responses, we pass the sixty-four prompts to a single-level GPT-3.5 Turbo model via API. 

\paragraph{Evaluation Metrics} 
All responses are individually provided to a local (offline) keep performance indicator evaluation agent. This agent is capable of both reference-based hallucination measurement through traditional NLP metric computations, as well as reference-free hallucination measurement (leveraging a custom fine-tuned version of Llama 3.1 70B). This agent produces a slate of 12 scores per response. 
For reach domain, we compute an aggregate score for each metrics, which is provided in Table \ref{tab:gt_llm_split_metrics}.

\section{Benchmark Results}

\begin{table*}[ht]
\centering
\caption{Average Ground Truth (GT) and LLM Hallucination Metrics for Politics \& Medicine (Split View)}
\begin{tabular}{llcccccc}
\toprule
\textbf{Domain} & \textbf{Source} & \textbf{FCD}$^{+}$ & \textbf{FGR}$^{+}$ & \textbf{FDF}$^{-}$ & \textbf{ECS}$^{-}$ & \textbf{SCD}$^{-}$ & \textbf{NRR}$^{+}$ \\
\midrule
\multirow{2}{*}{Politics}
  & GT  & 10.71 & 1.54 & 0.00 & 0.00 & 0.0000 & 0.0000 \\
  & LLM & 11.60 & 0.57 & 0.00 & 0.00 & 0.0441 & 0.0000 \\
\midrule
  &      & \textbf{CTC}$^{+}$ & \textbf{RCR}$^{+}$ & \textbf{ROUGE-L}$^{+}$ & \textbf{BLEU}$^{+}$ & \textbf{METEOR}$^{+}$ & \textbf{THS}$^{-}$ \\
\midrule
  & GT  & 0.0000 & 0.1486 & 1.0000 & 0.9218 & 0.9998 & 9.1786 \\
  & LLM & 0.0000 & 0.8250 & 0.1568 & 0.0096 & 0.1385 & 11.0307 \\
\midrule\midrule
\textbf{Domain} & \textbf{Source} & \textbf{FCD}$^{+}$ & \textbf{FGR}$^{+}$ & \textbf{FDF${^-}$} & \textbf{ECS$^{-}$} & \textbf{SCD}$^{-}$ & \textbf{NRR}$^{+}$ \\
\midrule
\multirow{2}{*}{Medicine}
  & GT  & 14.93 & 0.00 & 0.00 & 0.00 & 5.2517 & 0.0000 \\
  & LLM & 4.97  & 1.78 & 0.00 & 0.00 & 0.7064 & 0.0000 \\
\midrule
  &      & \textbf{CTC}$^{+}$ & \textbf{RCR}$^{+}$ & \textbf{ROUGE-L}$^{+}$ & \textbf{BLEU}$^{+}$ & \textbf{METEOR}$^{+}$ & \textbf{THS}$^{-}$ \\
\midrule
  & GT  & 0.0000 & 0.3148 & 1.0000 & 0.9514 & 0.9978 & 14.9292 \\
  & LLM & 0.0000 & 0.2858 & 0.0502 & 0.0030 & 0.1317 & 3.1875 \\
\bottomrule
\end{tabular}
\label{tab:gt_llm_split_metrics}
\end{table*}

\paragraph{Evidence Cited in Responses} The reported metrics suggest that there is a high hallucination potential for political prompts and a relatively lower potential for medical prompts. However, as we will discuss, the LLM’s seemingly superior performance in medicine compared to the ground truth likely reflects professional norms among physicians, who typically communicate diagnoses without citation or explicit justification.

In the political example, the Factual Claim Density (FCD$^{+}$) score is higher for the LLM than for the GT, and the Factual Grounding References (FGR$^{+}$) is lower in the LLM compared to the GT.\footnote{We present twelve metrics in this section. Fuller definitions of these metrics are included in \autoref{tab:metricdefinitions}, but to aid interpretation we mark each acronym with a superscript $^{+}$ or $^{-}$. Metrics marked with a plus (e.g. FCD$^{+}$) have less hallucination when they have higher values (i.e. higher is ``better''). Metrics marked with a minus (e.g. FDF$^{-}$) have less hallucination when they have lower values (i.e. lower is ``better'').} The interpretation is that more statements could be construed of as real facts (FCD$^{+}$) in the LLM compared to the GT, and that the LLM language appears to be less grounded in evidence (FGR$^{+}$) compared to the GT. In the medical example, the Factual Claim Density (FCD$^{+}$) score is lower for the LLM than for the GT, and the Factual Grounding References (FGR$^{+}$) is higher in the LLM compared to the GT. The interpretation is that fewer statements could be construed of as a real facts (FCD$^{+}$) in the LLM compared to the GT, and that the LLM language may appear to be more grounded in evidence compared to the GT (FGR$^{+}$). 

Our assessment is that the physician’s responses, reflective of how he communicates with patients in real life, did not cite sources or explicitly reference underlying medical theory. By comparison, the LLM responses often grounded their claims in recognizable medical knowledge, referencing specific conditions, medications, or treatment protocols. When we followed up with the physician about this, he explained that doctors do not cite their sources in conversations with patients. This approach contrasts with how a political practitioner might cite evidence or policy when addressing questions about procedures \citep[see, e.g.][]{ncsl2025election}. When we pressed the physician to think about what sources he would cite if he \textit{had to}, he replied: ``med school.'' When pushed further, the physician explained that justifications for diagnoses are taught and memorized in medical school, then internalized over years of clinical practice.

This example highlights a deeper issue: expert answers, while grounded in years of training and experience, may appear less grounded-in-truth than the LLM model responses. Without thoughtful prompt design and clear guidelines for what it means for experts to ``cite a source,'' hallucination benchmarks risk penalizing expert-informed answers while rewarding seemingly evidence-based responses from LLMs.

\paragraph{Structure of Responses} Our findings underscore the importance of aligning evaluation metrics with the specific task at hand. Several metrics --- Fictional Disclaimer Frequency (FDF${^-}$), Explicit Contextualization Score (ECS$^{-}$), Named Reference Ratio (NRR$^{+}$), Contextual Trigger Count (CTC$^{+}$) --- scored $0.0$ for both GT and LLMs in both domains. This is particularly significant: it implies that neither the GT responses nor the LLM generations employed structured disclaimers, explicit contextualization, or named references. This absence is not likely attributable to model hallucination; instead it is potentially due to the nature of the measurement task itself. The sample prompts and reference responses often consist of straightforward factual statements that lack the stylistic or rhetorical cues these metrics are designed to detect. These bottom-coded hallucination metrics highlight a fundamental limitation in applying conventional fictionality-sensitive metrics to question and answer domains that are firmly grounded in fact rather than fiction. 

\paragraph{Realism and Speculative Language} The text generated by the LLM mimics language that reflects greater realism or plausibility, but also includes language that is speculative. In the political example, the LLM responses averaged higher on the Real-world Concept Ratio (RCR$^{+}$) than the GT. This suggests that the LLM invoked more real-world concepts than the GT. However, this may not correspond to accurate grounding. The Speculative Claim Density (SCD$^{-}$) is higher for the LLM responses than for the GT, indicating an increased tendency to include language that hedges or speculates. The LLM produces text that has a higher THS$^{-}$ than GT, driven in large part by the relatively elevated FCD$^{+}$ and fewer contextual safeguards compared to the GT. The GT responses, in contrast, are shorter, more constrained, and structurally conservative---resulting in a low total hallucination score (THS$^{-}$), but also exhibit limited contextual sophistication (FGR$^{+}$, FDF${^-}$, and ECS$^{-}$). 

The board-certified, human-doctor produced responses that score \textit{considerably} higher on THS$^{-}$ than the model-generated text, principally due to the FCD$^{+}$ chasm. A naive interpretation of THS$^{-}$ would conclude that the doctor's responses presented a greater risk of being hallucinated than the model-generated text. Obviously, this cannot be true. In fact, the way that doctors think, reason, and communicate about diagnoses from chief-complaints is simply poorly measured by these hallucination metrics. As doctors professionalize and develop medical expertise they develop heuristics that allow them to produce correct diagnoses quickly. Given limited, and fallible cognition, they prioritize getting answers correct, rather than citing the grounding scientific study that connects a complaint to a diagnoses. And, when faced with chief-complaints, doctors hold in mind a series of possible diagnoses---leading to higher values of SCD$^{-}$---some of which would be ruled out pending the result of a follow-up laboratory tests. For example, when responding to patients' chief complaints and symptoms, doctors often say, ``It's possible you have {some diagnosis}. Let’s run some tests.''

\paragraph{Lexical Structure} Consistent with our expectations, LLM outputs appear fluent, but they often deviate in surface form and phrasing from the human-generated reference text. Such findings underscore the limitations of current LLM generations in matching not just factual content, but precise expression and structure. Across both the political and medical domains, metrics that evaluate similarity between generated text and what a human would have written (i.e. ROUGE-L$^{+}$, BLEU$^{+}$ and METEOR$^{+}$), GT responses score as near perfect matches to human written responses.  In contrast, LLM responses receive considerably lower scores; there is substantial lexical and structural differences between model-generated text and human-written, ground-truth responses. 

\section{Discussion}

One foreseeable challenge with any deployed language model system is that it will inevitably be used to perform expert-level tasks beyond what it was trained for. While we do not immediately propose a mechanism to solve this problem of boundless domain expertise, we suggest that making the testing and evaluation open---rather than closed---sourced would allow expertise encoded into models to be shared across teams. 
Our goal is to highlight the risks of using models for expert-level tasks without corresponding expert-level knowledge encoded within them.

A related and equally important challenge concerns the frequent disagreement among experts themselves. Especially near the knowledge frontier, the truth of a position may remain unsettled; in some domains, fundamental disagreements may never be resolved. As a consequence, querying one human expert rather than another to contribute training data could shape the responses generated by the resulting model. To represent this uncertainty, our preferred approach asks experts to first to articulate their own position and justification, and second to identify credible alternative positions and explain why they reject them. At a minimum, this provides the model with competing views that enable it to generate responses that better reflect the range of perspectives within a domain. 

Beyond disagreement, a perceive challenge is that using domain-expert assessments is prohibitively expensive, perhaps outweighing the benefits of reducing hallucinations. Expertise is costly---spot contracts for expert medical, legal, or scientific consulting often reach several hundred dollars per hour---and it is highly domain specific. Assessments and benchmarking tasks developed by experts in one field are unlikely to be useful even in closely-related, cognate domains. Nevertheless, these costs are minuscule compared to the enormous expense of training current-generation models and paying teams of data scientists and engineers. Moreover, model training produces model weights that quickly obsolesce and remain private goods: each generation of models is inevitably supplanted by the next, and model weights are rarely shared between companies or research teams. By comparison, domain-expertise is nonrivalrous---its use by one team does not diminish its availability to others---and it is also evergreen. Expert knowledge can be reused to benchmark successive generations of models and must only be updated when the underlying science, law, or art evolves. The implication is that costs to gather expertise can be amortized across a long time horizon. When shared within an open scientific community and distributed across many research groups, this expertise can serve as a public good that advances knowledge collectively. 

When expertise is treated as a shared public good, collective progress becomes possible---but this vision stands in contrast to the closed, competitive market dynamics that drive firms to develop ever more performant language models. One might argue that the commercial landscape is better positioned to invest in benchmarks than the open scientific community. While it is true that commercially developed language models have dominated public attention, it is important to remember that the theoretical foundations for these models were established through open scientific dialogue \citep[e.g.,][]{vaswani2017attention}. Commercial firms' advantage lies in their ability to mobilize resources to glean text, process it through compute to create proprietary model weights, and deliver packaged product to consumers. Benchmarking, by contrast, benefits from coordination rather than competition. Benchmarks are non-rivalrous and most effective when shared within and across teams; moreover, they are not customer-facing products and thus do not require polished front-ends. For precisely these reasons, market competition is likely to under-provide open, domain-contextualized benchmark data, since such data can be freely used by competitors to improve their models. 

Looking forward, there are reasons to believe that the same collaborative dynamics that led to the success of the open-source software paradigm could also drive progress in expert benchmarking. Companies derive value from the applications built on top of core tools, and coordination around a common toolset (or benchmark set) enables collaboration within and between firms without revealing trade secrets. Like open-source development, companies that participate in open benchmarking gain access to a broader talent pool of researchers already familiar with widely used frameworks, reducing the need for costly internal training on proprietary tooling. Moreover, benchmark maintenance need not rest with a single organization but can instead be distributed among domain experts in the global research community.

\section{Conclusion}
In this paper, we argue that subjecting language models to repeatable, open, and domain-contextualized hallucination benchmarks will advance language model development. Repeatability is essential for internal testing and scientific rigor because tuning models to perform well on narrow external benchmarks often leads to overfitting: strong performance on the benchmark task but poor generalization to other contexts. Openness in benchmark development is foundational to science and enables users to independently verify claims made by software providers. Moreover, open benchmarks foster collaboration across organizations and between academic and industry groups, facilitate cost-sharing when procuring questions and answers, and invite contributions to be volunteered from domain experts, ensuring that benchmarks reflect current knowledge. Domain-contextualized benchmarks that represent expert thinking are necessary for hallucination metrics to remain effective and relevant as models are applied to increasingly complex problems. Failing to incorporate such expertise risks deploying language models that are not only unhelpful but potentially harmful.

\clearpage
\bibliographystyle{abbrv}
\bibliography{references}

\clearpage

\begin{sidewaystable}
    \centering
    \vspace*{7cm}
    \begin{tabular}{lp{4cm}p{7cm}p{7cm}}
    \toprule 
     Abbrev. & Metric & Definition & Interpretation \\
     \midrule
     FCD$^{+}$ & Factual Claim Density & Number of claims that appear to be historical, scientific, or verifiable facts per 100 words. & A lower FCD$^{+}$ suggests fewer statements that could be mistaken for real facts. \\
     FGR$^{+}$ & Factual Grounding References & Number of references to \textit{real-world} evidence & Lower FGR$^{+}$ texts are more overtly fictional. \\ 
     FDF${^-}$ & Fictional Disclaimer Frequency & Frequency (per 100 tokens) of explicit indications the text is fictional. & Higher FDF${^-}$ means texts are framed as not-based-in-fact. \\ 
     ECS$^{-}$ & Explicit Contextualization Score & Frequency (per 100 tokens) text includes language such as ``purely fictional'' or ``mythical.'' & Higher ECS$^{-}$ indicates stronger distancing from factual framing. \\ 
     \textit{THS}$^{-}$ & Total Hallucination Score & Composite: FCD$^{+}$ - (FGR$^{+}$ + FDF${^-}$ + ECS$^{-}$). & Higher THS$^{-}$ implies a denser presence of fact-like claims that are not offset by disclaimers or real-world grounding. \\
     SCD$^{-}$ & Speculative Claim Density & Proportion of sentences containing hedging or speculative language (e.g., ``might,'' ``some believe,'' ``possibly''). & Higher SCD$^{-}$ may reflect caution in LLMs, but also signals epistemic uncertainty that can blur fact vs. fiction boundaries. \\
     NRR$^{+}$ & Named Reference Ratio & Proportion of sentences that include named entities tied to identifiable sources (e.g., people, organizations, documents). & Higher NRR$^{+}$ suggests stronger citation patterns; low values often signal decontextualized or imagined references. \\
     CTC$^{+}$ & Contextual Trigger Count & Number of explicit framing cues per 100 tokens (e.g., ``according to legend,'' ``myth says,'' ``reportedly''). & Higher CTC$^{+}$ indicates the model is flagging uncertainty or fictionality more clearly. \\
     RCR$^{+}$ & Real-world Concept Ratio & Proportion of noun phrases or terms that refer to concrete, verifiable real-world entities. & Higher RCR$^{+}$ reflects greater realism or plausibility; low values suggest fictional abstraction or vagueness. \\
     ROUGE-L$^{+}$ & Recall-Oriented Understudy for Gisting Evaluation & Measures the longest common subsequence between generated and reference text. & Higher ROUGE-L$^{+}$ indicates stronger structural and lexical alignment with the reference. \\ 
     BLEU$^{+}$ & Bilingual Evaluation Understudy & Precision-oriented score capturing overlap between n-grams in the candidate and reference. & High BLEU$^{+}$ reflects lexical fidelity to the reference; low BLEU$^{+}$ suggests phrasing divergence. \\ 
     METEOR$^{+}$ & Metric for Evaluation of Translation with Explicit ORdering & Harmonic mean of unigram precision and recall, with penalties for word order and semantic mismatch. & More tolerant than BLEU$^{+}$; higher METEOR$^{+}$ values indicate better fluency and synonym-aware alignment with the reference. \\
    \bottomrule
    \end{tabular}
    \caption{FCD$^{+}$, FGR$^{+}$, FDF${^-}$, ECS$^{-}$, and the composite THS$^{-}$ are drawn from Gosmar \& Dahl, 2025 \citep{gosmar2025}. Additional reference-free metrics (SCD$^{-}$, NRR$^{+}$, CTC$^{+}$, RCR$^{+}$) \citep{lin2022teaching,wadden2021multivers,zhao2023felm,petroni2019language} extend interpretability in contexts lacking ground-truth. ROUGE-L$^{+}$ \citep{lin2004}, BLEU$^{+}$ \citep{papineni2002}, and METEOR$^{+}$ \citep{banerjee2005} are reference-based metrics standard in natural language evaluation.}
    \label{tab:metricdefinitions}
\end{sidewaystable}

\end{document}